# Wind speed prediction using a hybrid model of the multi-layer perceptron and whale optimization algorithm

Saeed Samadianfard [1], Sajjad Hashemi [1], Katayoun Kargar [2], Mojtaba Izadyar [1], Ali Mostafaeipour [3], Amir Mosavi [4*] Narjes Nabipour [5*], Shahaboddin Shamshirband [6,7*]

[1] Department of Water Engineering, Faculty of Agriculture, University of Tabriz, Tabriz, Iran, s.samadian@tabrizu.ac.ir
[2] Department of Civil Engineering, Faculty of Engineering, Urmia University, Urmia, Iran,
[3] Industrial engineering Department, Yazd University, Yazd, Iran, mostafaei@yazd.ac.ir
[4] Kalman Kando Faculty of Electrical Engineering, Obuda University, 1034 Budapest, Hungary; amir.mosavi@kvk.uni-obuda.hu
[5] Institute of Research and Development, Duy Tan University, Da Nang 550000, Viet Nam Corresponding authors. narjesnabipour@duytan.edu.vn
[6] Department for Management of Science and Technology Development, Ton Duc Thang University, Ho Chi Minh City, Viet Nam; shahaboddin.shamshirband@tdtu.edu.vn,
[7] Faculty of Information Technology, Ton Duc Thang University, Ho Chi Minh City, Viet Nam

**Abstract**

Wind power as a renewable source of energy, has numerous economic, environmental and social benefits. In order to enhance and control the renewable wind power, it is vital to utilize models that predict wind speed with high accuracy. Due to neglecting of requirement and significance of data preprocessing and disregarding the inadequacy of using a single predicting model, many traditional models have poor performance in wind speed prediction. In the current study, for predicting wind speed at target stations in the north of Iran, the combination of a multi-layer perceptron model (MLP) with the Whale Optimization Algorithm (WOA) used to build new method (MLP-WOA) with a limited set of data (2004-2014). Then, the MLP-WOA model was utilized at each of the ten target stations, with the nine stations for training and tenth station for testing (namely: Astara, Bandar-E-Anzali, Rasht, Manjil, Jirandeh, Talesh, Kiyashahr, Lahijan, Masuleh and Deylaman) to increase the accuracy of the subsequent hybrid model. Capability of the hybrid model in wind speed forecasting at each target station was compared with the MLP model without the WOA optimizer. To determine definite results, numerous statistical performances were utilized. For all ten target stations, the MLP-WOA model had precise outcomes than the standalone MLP model. The hybrid model had acceptable performances with lower amounts of the RMSE, SI and RE parameters and higher values of NSE, WI and KGE parameters. It was concluded that WOA optimization algorithm can improve prediction accuracy of MLP model and may be recommended for accurate wind speed prediction.

## 1. Introduction

By increasing the need for energy in today's societies and declining fossil resources, the importance of renewable energies appears more than ever. Wind energy, as a substitute of fossil resources, has received rising attention from all over the world owing to its abundant supply, extensive dispersal, and finances as a clean and renewable form of energy. Also, rising alertness of the ecological effects of greenhouse gas releases has encouraged an impressive rise in renewable energy. Therefore, to encounter the energy request and the problems of greenhouse gas releases, it is essential to concentrate on substitute renewable energies (Deo et al. 2018, hoolohan et al. 2018, and Marchal et al. 2011). Although the wind supply in most parts of the world is plentiful, its unpredictable and irregular nature lead to some problems such as acquiring a safe and persistent supply of electricity. By predicting the wind power, the request for electricity can be cautiously controlled and their precision has a direct effect on consistency and productivity (hoolohan et al. 2018).

Local and regional climates, topography, and impediments including buildings affect wind energy. Due to the cyclical, daily pattern and high stochastic variability, accurate prediction of wind power is too complicated. Therefore, it is clear that efficient transformation and application of the wind energy resources require exact and complete information on the wind features of the region. Wind power prediction relies on wind speed estimation. In the last decades, different models was established to predict the wind speed to reach accurate information about wind energy. In general, these models are divided into three types: physical, statistical, and intelligence learning models. Physical approaches which are based on a detailed physical description of the atmosphere, used meteorological data such as air temperature, topography, and pressure to predict wind speed. These type of methods have not been applied in short-term wind speed prediction owing to intricate calculation methods, high costs, and poor performance, but they can have more accurate predictions in long-term compared with other types of prediction models. For example, Cheng et al. (2017) used physical algorithms to integrate observation data of wind turbines into numerical weather prediction (NWP) systems to enhance the precision of wind speed forecasting. Moreover, Charabi et al. (2011) and Al-Yahyai and Charabi (2015) evaluated wind sources in Oman by NWP models, and Jiang et al. (2013) investigated wind energy capacities in coastal regions of china by utilization of remotely sensed wind field information. For short-term periods statistical methods and intelligence learning models, which have been applied in most of the recent studies, can forecast wind speed better and more accurate than physical approaches. The autoregressive (AR), autoregressive moving average (ARMA), and the autoregressive integrated moving average (ARIMA) models are used as statistical methods. As an example of statistical methods, Torres et al. (2005) predicted wind speeds up to 10 hours earlier by applying the ARMA model in Navarre (Spain). Enhancements over a persistence model were presented in the study, but it was noted that the model could only be used in short-term predictions. Kavasseri Rajesh and Seetharaman Krithika (2009) utilized the fractional autoregressive integrated moving average (f-ARIMA) model to predicted wind speed for upcoming two-day periods. The results expressed that the precision of f-ARIMA model was higher than the persistence model. In the case of intelligence learning models, fuzzy systems, artificial neural networks (ANN), support vector regression (SVR), neuro-fuzzy systems, extreme learning machines, and the Gaussian process are the most current methods for wind prediction. Also, hybrid models are used for wind speed forecasting, which are usually made with the combination of statistical and intelligent methods (chitsazan et al. 2019). Shukur and Lee (2015) used the data from Malaysia and Iraq in order to predict daily wind speed by the utilization of a hybrid model with a combination of an artificial neural network (ANN) and Kalman filter (KF). The outcomes showed that the KF-ANN as a hybrid model had high performance in comparison with single algorithms. Bilgili and Sahin (2013) predicted wind speed in daily, weekly, and monthly periods by exploiting the ANN method with data from four different stations of Turkey. The results showed that the applied method performed well. Moreno and Coelho (2018) exploited the Adaptive Neuro-Fuzzy Inference System (ANFIS) with a combination of Singular Spectrum Analysis (SSA) for wind speed predicting. The results expressed that forecasting errors were considerably reduced by the utilization of the proposed method. Cadenas and Rivera (2010) developed hybrid models, including ANN and ARIMA models to forecast wind speed in three different locations. First, they used the ARIMA model to forecast wind speed of time sequences, and the ANN model was used to considering the nonlinear features that the ARIMA model could not recognize. It was concluded that in this process, the hybrid models are more precise than the ANN and ARIMA models. Hui Liu et al. (2015) integrated four decomposing algorithms including Empirical Mode Decomposition (EMD), Fast Ensemble Empirical Mode Decomposition (FEEMD), Wavelet Decomposition (WD), and Wavelet Packet Decomposition (WPD) with two nominated networks including ANFIS and MLP Neural Network to estimate wind speed. Based on the results, the hybrid ANN algorithms have high accuracy in comparison with their corresponding single ANN algorithms in wind speed prediction. Furthermore, the ANFIS had poor performance than the MLP in the forecasting neural networks.

In this study, a hybrid technique was developed based on an MLP model for predicting the wind speed without any requirement to the atmospheric datasets. Therefore, to predict the wind speed value of the target station, data of reference stations were used. Moreover, to improve the precision of the model, the whale optimization algorithm (WOA) is utilized and novel MLP-WOA model is developed. The WOA model has been used as an optimizer in earlier investigations (e.g. Du et al. 2018) in electrical power forecasting, but the aim of this research is investigating of MLP-WOA model for wind speed forecasting for a set of ten spatially-scattered stations in the north of Iran by applying data of the reference stations.

This paper is structured as follows: In the next section, the methods and materials are described in detail. The results and discussions of the models are presented in section 3, and lastly, section 6 presents the conclusions.

## 2. Methods and materials

### 2.1. Multilayer perceptron neural networks

Multilayer perceptron models, which are constructed based on nervous system of human brain, has high capabilities in modeling nonlinear behavior of complex systems. Furthermore, the nature of these models allows them to address prediction problems with nonlinear structure. This model operates on the basis of learning the problem-solving process for reaching the output by finding the implicit relationship in the process. For this purpose, a bunch of data is used in the training stage, by the usage of the relationship found in that stage, then, the proper output is calculated. There are several samples of the neural networks but among all of them, the back-propagation network is used more than others. This network consists of layers and they have parallel-acting elements called neurons. Each layer is entirely connected to layer before and after itself.

In this study, the composition of (i) input layer, (ii) hidden layer, and (iii) output layer is used as a three-layered structure (Figure 1). The independent parameters in the input layer consist of nine neighboring stations. The dependent variable that utilized as an output is the target station. The optimum network design includes 9, 8 and neurons for input, hidden and output layers, respectively. Moreover, the sigmoid tangent and linear functions using the Lewenberg Marquard Algorithm (LMA) with 200 repeating were utilized for input and output layers.

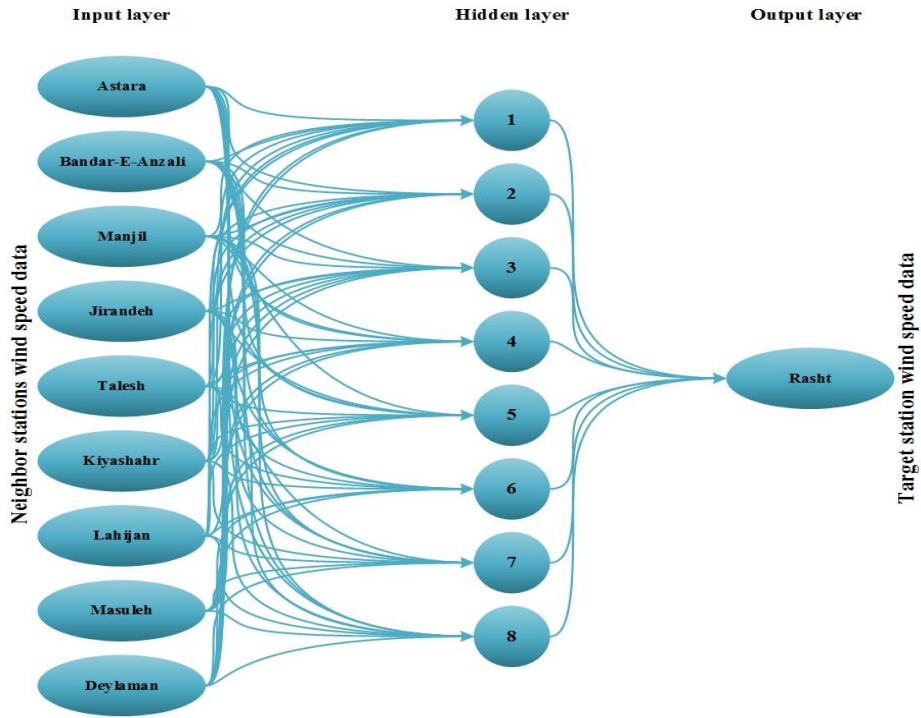

Figure 1. Artificial neural network arrangement in this study

## 2.2. Multi-layer Perceptron-Whale Optimization Algorithm (MLP-WOA)

Mirjalili and Lewis (2016) suggested whale optimization algorithm which is a new heuristic algorithm. WOA impersonators the foraging of humpback whales. The humpback whales have particular hunting method identified as a bubble-net feeding technique in which they catch a group of small fishes near the surface. They create distinctive bubbles along a spiral-shaped rout by swimming around prey within a diminishing circle (Fig. 2). The WOA is done in two stages. The first one is exploitation in which the prey is encircled and the bubble spiral attack technique is used, and in the second step, prey selected randomly which is named exploration.

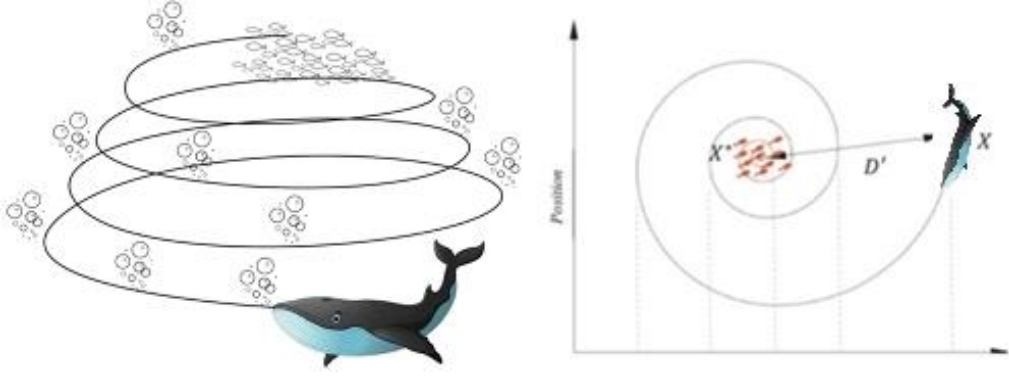

Figure 2. Artificial neural network-whale optimization algorithm (MLP-WOA)

The WOA can discover the situation of the hunt to encircle them. In the whale method, it is supposed that the present best location is target prey or it is near the optimum since the optimum search location is not defined earlier. The following equations characterize this performance:

$$\vec{D} = |\vec{C}.\vec{X^*} - \vec{X}(t)| \tag{1}$$

$$\vec{X}(t+1) = \vec{X^*}(t) - \vec{A}.\vec{D} \tag{2}$$

Where $\vec{C}$ and $\vec{A}$ are considered as coefficient vectors, $t$ represents the current iteration, $\vec{X}$ is the location vector *and* $X^*$ is the location vector of the best solution. The following equations represent A and C:

$$\vec{A} = 2\vec{a}.\vec{r} - \vec{a} \tag{3}$$

$$\vec{C} = 2.\vec{r} \tag{4}$$

where $r$ is a random vector produced with steady diffusion in the interval of [0, 1] and $a$ declines from two to zero by order of iterations. In Eq. (2) solutions verify their locations according to the site of the best solutions (prey). In WOA for achieving the shrinking encircling behavior in a trap, $a$ is reduced with the subsequent formula:

$$a = 2 - t\frac{2}{MaxIter} \tag{5}$$

where $t$ is repeating number and MaxIter is the maximum allowable iterations. The distance between the best known search ($X^*$) and a search factor (X) is calculated to simulate the spiral-

shaped route. Then to create the adjacent search agent location, a spiral equation is formed as follows:

$$\vec{X}(t+1) = D'.e^{bl}.\cos(2\pi L) + \vec{X^*}(t) \qquad (6)$$

where $L$ is a random number in $[-1,1]$, $b$ is a constant and the space of the $i$th whale and the prey is considered as $D'$ which is calculated by:

$$D' = \left|\vec{X^*}(t) - \vec{X}(t)\right| \qquad (7)$$

As mentioned above, the humpback whales swimming around preys in a diminishing circular as well as a spiral-shaped route simultaneously. To simulate the two mechanisms, during the optimization process there is a likelihood of 50% to select between them:

$$\vec{X}(t+1) = \begin{cases} Shrinking\ Encircling\ (eq.5) & (P < 0.5) \\ Spiral - shaped\ path\ (eq.9) & (P \geq 0.5) \end{cases} \qquad (8)$$

where $P$ is a random number in [0, 1]. In the current research, the value of $L$ and P were 0.65 and 0.37, respectively. Also the size of population was 30 and maximum iteration was 50. Furthermore, the optimum number of neurons was considered 8 in the hidden layer.

## 2.3 Accuracy appraisal Standards

Several statistical parameters have been utilized to measure the accuracy of the models. In the present study, various statistical parameters, including Determination coefficient ($R^2$), Root mean square errors (RMSE), Present relative error (RE), Willmott's Index (WI), Scatter Index (SI), Nash-Sutcliffe efficiency (NSE), and Kling-Gupta efficiency (KGE) are utilized. These accuracy criteria are defined as follows.

$$R^2 = \left[\frac{\left(\sum_{i=1}^{n} O_i P_i - \frac{1}{n}\sum_{i=1}^{n} O_i \sum_{i=1}^{n} P_i\right)}{\left[\sum_{i=1}^{n} O_i^2 - \frac{1}{n}\left(\sum_{i=1}^{n} O_i\right)^2\right]\left[\sum_{i=1}^{n} P_i^2 - \frac{1}{n}\left(\sum_{i=1}^{n} P_i\right)^2\right]}\right]^2 \qquad (9)$$

$$RMSE = \sqrt{\frac{1}{n}\sum_{i=1}^{n}(P_i - O_i)^2} \qquad (10)$$

$$WI = 1 - \left[ \frac{\sum_{i=1}^{n}(O_i - P_i)^2}{\sum_{i=1}^{n}\left(\left|P_i - \overline{O}_i\right| + \left|O_i - \overline{O}_i\right|\right)^2} \right] \quad (11)$$

$$SI = \frac{\sqrt{\frac{1}{n}\sum_{i=1}^{n}(P_i - O_i)^2}}{\overline{O}} \quad (12)$$

$$NSE = 1 - \frac{\sum_{i=1}^{n}(P_i - O_i)^2}{\sum_{i=1}^{n}(O_i - \overline{O})^2} \quad (13)$$

$$KGE = 1 - \sqrt{(r-1)^2 + (\beta-1)^2 + (\gamma-1)^2} \quad (14)$$

$$r = \frac{\sum_{i=1}^{n}(O_i - \overline{O})(P_i - \overline{P})}{\sqrt{\sum_{i=1}^{n}(O_i - \overline{O})^2 \sum_{i=1}^{n}(P_i - \overline{P})^2}} \qquad \beta = \frac{\overline{P}}{\overline{O}} \qquad \gamma = \frac{CV_P}{CV_o}$$

Where n is the number of data set. $O_i$ and $P_i$ are the observed and estimated values. Also, $\sigma_o$ and $\sigma_p$ is the standard deviation of observed and estimated values from MLP or MLP-WOA, individually.

**2.4. Study area and predictive model development**

In the present study, the monthly mean wind speed data of ten locations in Gilan province, over the period of 2004-2014 were collected. The studied stations included: Astara, Bandar-E-Anzali, Rasht, Manjil, Jirandeh, Talesh, Kiyashahr, Lahijan, Masuleh, and Deylaman (Fig.4). Latitude and longitude of studied stations vary between 36º42′ to 38º21′ North and 48º51′ to 50º00′ East respectively, while their height above sea level differs between -23.6 and 1581.4 m a.s.l. Table 1 shows coordinates of studied stations in the region and the statistical characteristics of wind data. Relative to the other stations, the lowest mean wind speed belongs to the Lahijan station (≈ 1.46 ms$^{-1}$), whereas the station with the windiest climate is Jirandeh with the mean wind speed of 5.25 ms$^{-1}$. Furthermore, Jirandeh station with the value of 25.6 ms$^{-1}$ had the maximum wind speed in the studied period.

Table 2 presented the list of reference and target stations in the studied region. Also, the correlation values of wind speed between target and reference stations are shown at Table 3.

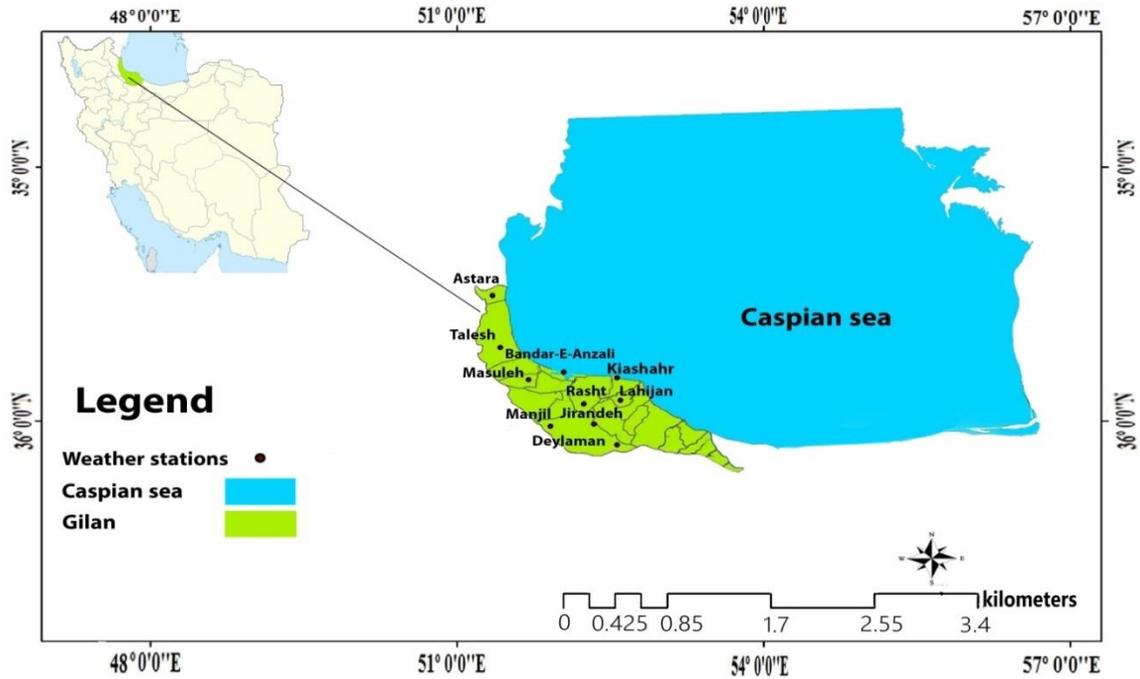

**Fig. 4.** The location of studied stations in the region

**Table. 1** Coordinates of studied stations in the region and the statistical characteristics of wind data.

| Station | Latitude | Longitude | Altitude(m) | Mean Wind speed(m/sec) | Maximum Wind Speed(m/sec) |
|---|---|---|---|---|---|
| Astara | 38º 21' 53.9"N | 48º 51' 17.6"E | -21.1 | 1.48 | 12.8 |
| Bandar-E-Anzali | 37º 28' 46.6"N | 49º 27' 27.2"E | -23.6 | 3.31 | 14.6 |
| Rasht | 37º 19' 21.9"N | 49º 37' 25.8"E | -8.6 | 1.51 | 9.0 |
| Manjil | 36º 43' 42.4"N | 49º 24' 36.0"E | 338.3 | 5.02 | 15.3 |
| Jirandeh | 36º 42' 27.5"N | 49º 48' 05.6"E | 1581.4 | 5.25 | 25.6 |
| Talesh | 37º 50' 22.5"N | 48º 53' 51.2"E | 7 | 1.75 | 17.8 |
| Kiyashahr | 37º 23' 21.0"N | 49º 53' 37.5"E | -22 | 1.66 | 10.0 |
| Lahijan | 37º 11' 32.5"N | 50º 00' 58.2"E | 34.2 | 1.46 | 10.6 |
| Masuleh | 37º 09' 02.3"N | 48º 59' 09.9"E | 1080.9 | 1.78 | 12.8 |
| Deylaman | 36º 53' 08.2"N | 49º 54' 35.7"E | 1447.6 | 2.38 | 14.6 |

Table. 2 Reference and target stations in the studied region.

| Target Station | Reference Station | Models | |
|---|---|---|---|
| Astara | Bandar-E-Anzali, Rasht, Manjil, Jirandeh, Talesh, Kiyashahr, Lahijan, Masuleh, Deylaman | MLP1 | MLP-WOA1 |
| Bandar-E-Anzali | Astara, Rasht, Manjil, Jirandeh, Talesh, Kiyashahr, Lahijan, Masuleh, Deylaman | MLP2 | MLP-WOA2 |
| Rasht | Astara, Bandar-E-Anzali, Manjil, Jirandeh, Talesh, Kiyashahr, Lahijan, Masuleh, Deylaman | MLP3 | MLP-WOA3 |
| Manjil | Astara, Bandar-E-Anzali, Rasht, Jirandeh, Talesh, Kiyashahr, Lahijan, Masuleh, Deylaman | MLP4 | MLP-WOA4 |
| Jirandeh | Astara, Bandar-E-Anzali, Rasht, Manjil, Talesh, Kiyashahr, Lahijan, Masuleh, Deylaman | MLP5 | MLP-WOA5 |
| Talesh | Astara, Bandar-E-Anzali, Rasht, Manjil, Jirandeh, Kiyashahr, Lahijan, Masuleh, Deylaman | MLP6 | MLP-WOA6 |
| Kiyashahr | Astara, Bandar-E-Anzali, Rasht, Manjil, Jirandeh, Talesh, Lahijan, Masuleh, Deylaman | MLP7 | MLP-WOA7 |
| Lahijan | Astara, Bandar-E-Anzali, Rasht, Manjil, Jirandeh, Talesh, Kiyashahr, Masuleh, Deylaman | MLP8 | MLP-WOA8 |
| Masuleh | Astara, Bandar-E-Anzali, Rasht, Manjil, Jirandeh, Talesh, Kiyashahr, Lahijan, Deylaman | MLP9 | MLP-WOA9 |
| Deylaman | Astara, Bandar-E-Anzali, Rasht, Manjil, Jirandeh, Talesh, Kiyashahr, Lahijan, Masuleh | MLP10 | MLP-WOA10 |

Table. 3 Correlation coefficient values of wind speed among all studied stations two by two.

| Station | Astara | Bandar-E-Anzali | Rasht | Manjil | Jirandeh | Talesh | Kiyashahr | Lahijan | Masuleh | Deylaman |
|---|---|---|---|---|---|---|---|---|---|---|
| Astara | 1.00 | | | | | | | | | |
| Bandar-E-Anzali | 0.44 | 1.00 | | | | | | | | |
| Rasht | 0.48 | 0.71 | 1.00 | | | | | | | |
| Manjil | 0.20 | 0.31 | 0.27 | 1.00 | | | | | | |
| Jirandeh | 0.19 | 0.29 | 0.26 | 0.70 | 1.00 | | | | | |
| Talesh | 0.29 | 0.05 | 0.18 | 0.16 | 0.16 | 1.00 | | | | |
| Kiyashahr | 0.40 | 0.50 | 0.54 | 0.15 | 0.19 | 0.17 | 1.00 | | | |
| Lahijan | 0.38 | 0.37 | 0.46 | 0.15 | 0.12 | 0.28 | 0.40 | 1.00 | | |
| Masuleh | 0.07 | -0.18 | 0.00 | -0.28 | -0.09 | 0.16 | 0.15 | 0.06 | 1.00 | |
| Deylaman | 0.24 | 0.08 | 0.20 | 0.10 | 0.27 | 0.13 | 0.28 | 0.11 | 0.43 | 1.00 |

## 3. Result and discussion

In this research, the abilities of both the MLP model and MLP optimized model with WOA in predicting wind speed by using datasets of nine neighboring sites in the North of Iran were investigated and compared with each other. In this research, by the usage of nine adjoining stations, wind speed of the target station is estimated by two models of MLP and MLP-WOA. Moreover, there is no straightforward way of splitting training and testing data. For instance, the study of Kurup and Dudani (2014) utilized a total of 63% of their data for model development, whereas Qasem et al., (2019) utilized 67% of data and Deo et al. (2018), Samadianfard et al. (2018), and Samadianfard et al. (2019a,b) used 70% and Zounemat-Kermani et al., (2019) implemented 80% of entire data to develop their models. Consequently, to create models for wind speed prediction, 70% of the data (2534 data) is applied for training, and 30% of them (1077 data) is utilized for the testing phase. It should be noted that code was written in the Wolfram Mathematica software so that the dataset is randomly selected for each two training and testing period for several times. Then the desired model was selected based on the best values for the determination coefficient ($R^2$) and the root mean square error (RMSE). After 50 repetitions of the above-mentioned random selection criteria in the Wolfram Mathematica software, the best conditions for $R^2$ and RMSE were selected and the data was entered to the process of the WOA method (Fig3).

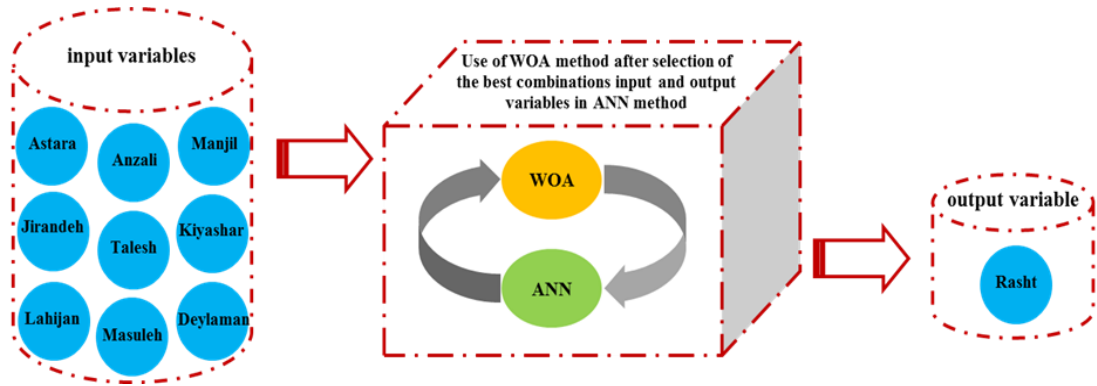

Figure 3. Proposed methodology for new hybrid model development

So, Table 4 shows statistical results of different MLP and WMLP models. Moreover, Fig. 5 shows bar graphs of the statistical parameters in testing phase.

**Table. 4** Statistical results of comparing different MLP and WMLP models.

| Models | Training | Testing |
|---|---|---|

|  | RMSE | SI | WI | NS | KGE | RMSE | SI | WI | NS | KGE |
|---|---|---|---|---|---|---|---|---|---|---|
| MLP1 | 0.789 | 0.519 | 0.741 | 0.378 | 0.506 | 0.723 | 0.522 | 0.725 | 0.346 | 0.454 |
| MLP2 | 1.013 | 0.308 | 0.899 | 0.683 | 0.767 | 1.185 | 0.353 | 0.885 | 0.624 | 0.758 |
| MLP3 | 0.532 | 0.362 | 0.889 | 0.656 | 0.715 | 0.623 | 0.385 | 0.856 | 0.620 | 0.656 |
| MLP4 | 1.907 | 0.386 | 0.900 | 0.685 | 0.761 | 2.424 | 0.470 | 0.832 | 0.561 | 0.603 |
| MLP5 | 2.953 | 0.547 | 0.865 | 0.608 | 0.670 | 2.995 | 0.615 | 0.837 | 0.507 | 0.649 |
| MLP6 | 0.862 | 0.491 | 0.697 | 0.353 | 0.405 | 0.784 | 0.447 | 0.592 | 0.197 | 0.286 |
| MLP7 | 0.868 | 0.509 | 0.775 | 0.443 | 0.524 | 0.570 | 0.367 | 0.819 | 0.335 | 0.676 |
| MLP8 | 0.942 | 0.686 | 0.745 | 0.402 | 0.446 | 0.814 | 0.483 | 0.689 | 0.092 | 0.493 |
| MLP9 | 1.284 | 0.633 | 0.767 | 0.438 | 0.526 | 1.184 | 0.957 | 0.661 | 0.224 | 0.206 |
| MLP10 | 1.356 | 0.509 | 0.789 | 0.442 | 0.586 | 0.938 | 0.549 | 0.727 | 0.016 | 0.493 |
| MLP-WOA1 | 0.729 | 0.479 | 0.787 | 0.469 | 0.554 | 0.657 | 0.474 | 0.771 | 0.461 | 0.518 |
| MLP-WOA2 | 0.900 | 0.274 | 0.921 | 0.750 | 0.796 | 1.078 | 0.321 | 0.913 | 0.589 | 0.779 |
| MLP-WOA3 | 0.473 | 0.322 | 0.913 | 0.728 | 0.746 | 0.523 | 0.323 | 0.908 | 0.732 | 0.705 |
| MLP-WOA4 | 1.695 | 0.343 | 0.922 | 0.751 | 0.789 | 2.086 | 0.405 | 0.887 | 0.675 | 0.654 |
| MLP-WOA5 | 2.747 | 0.509 | 0.886 | 0.660 | 0.694 | 2.751 | 0.565 | 0.870 | 0.584 | 0.679 |
| MLP-WOA6 | 0.801 | 0.457 | 0.750 | 0.440 | 0.457 | 0.703 | 0.401 | 0.689 | 0.354 | 0.387 |
| MLP-WOA7 | 0.810 | 0.475 | 0.813 | 0.516 | 0.561 | 0.548 | 0.353 | 0.841 | 0.386 | 0.707 |
| MLP-WOA8 | 0.880 | 0.641 | 0.788 | 0.478 | 0.485 | 0.722 | 0.429 | 0.753 | 0.285 | 0.585 |
| MLP-WOA9 | 1.190 | 0.587 | 0.808 | 0.517 | 0.568 | 1.097 | 0.887 | 0.732 | 0.334 | 0.255 |
| MLP-WOA10 | 1.206 | 0.452 | 0.836 | 0.559 | 0.645 | 0.903 | 0.529 | 0.766 | 0.058 | 0.524 |

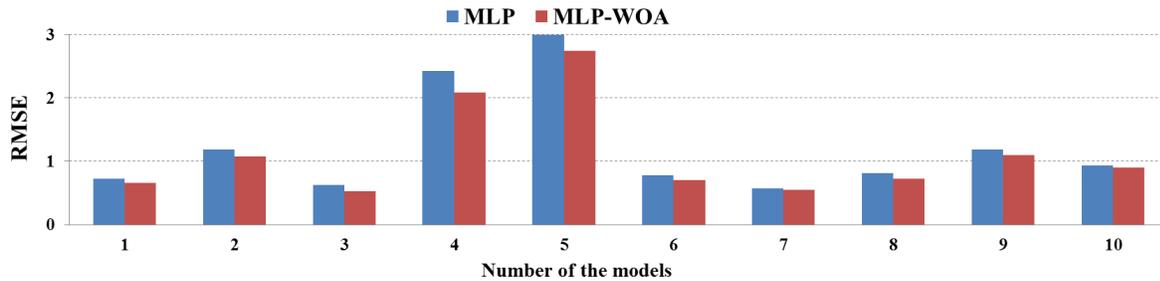

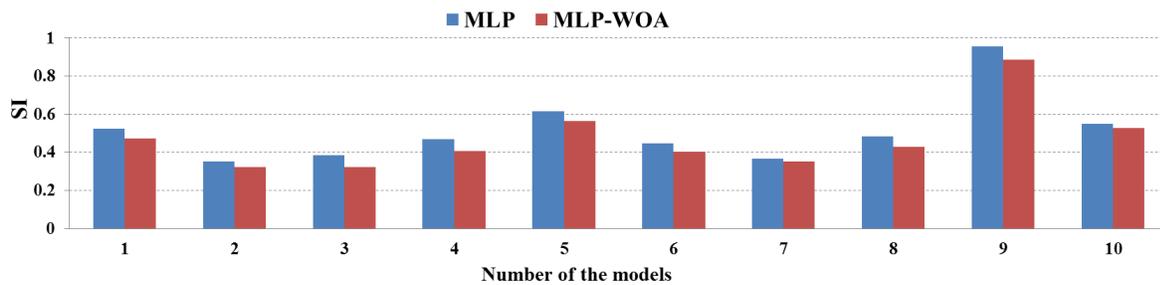

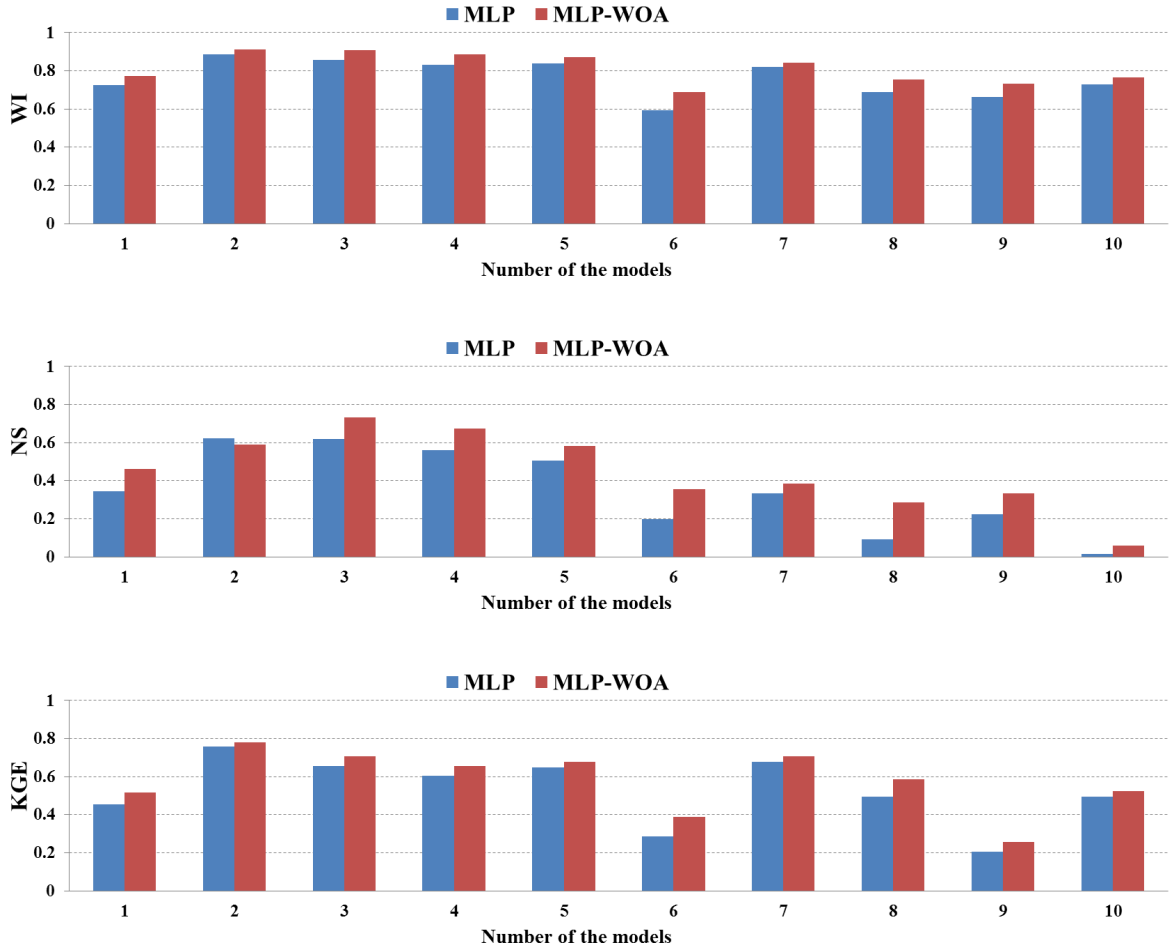

**Fig. 5.** Bar graphs of the statistical parameters for different considered models.

The RMSE and SI of models that optimized with WOA were lower than standalone MLP models at all stations in the training phase. Also, in the testing phase, MLP-WOA models had better performance than classical MLP models, and accuracy of optimized models were higher than standalone MLP models, so that RMSE of MLP models varied between 0.57 and 1.18. Whereas, for models optimized with WOA algorithm, this value was decreased to reach the range 0.52-1.09. Two stations of Manjil and Jirandeh in both MLP and MLP-WOA models had higher RMSE values. Between the studied stations, in classical MLP models, Kiyashahr and Rasht had the best performance with the RMSE of 0.57 and 0.62 and SI of 0.36 and 0.38, respectively. Similarly, in WOA-MLP models, the mentioned stations were the most accurate models with the RMSE of 0.54 and 0.52 and SI of 0.35 and 0.32, respectively. Moreover, according to other statistical parameters that used in this study, WI, NS, and KGE of the models that optimized with the WOA algorithm demonstrated better performance in comparison with classical MLP models. Also, RE of the models decreased after optimizing by the WOA algorithm. Fig. 6 demonstrated the performance of the hybrid MLP-WOA model in comparison with the standalone MLP model for ten study stations (Fig. 6). As mentioned, it can be concluded from Fig. 6 that the WOA algorithm improved the accuracy of winds peed forecasting of the MLP model. Moreover, to further evaluation of the precision of the developed models, a scatter plot of observed and predicted wind speed between the two datasets is presented in Fig. 7.

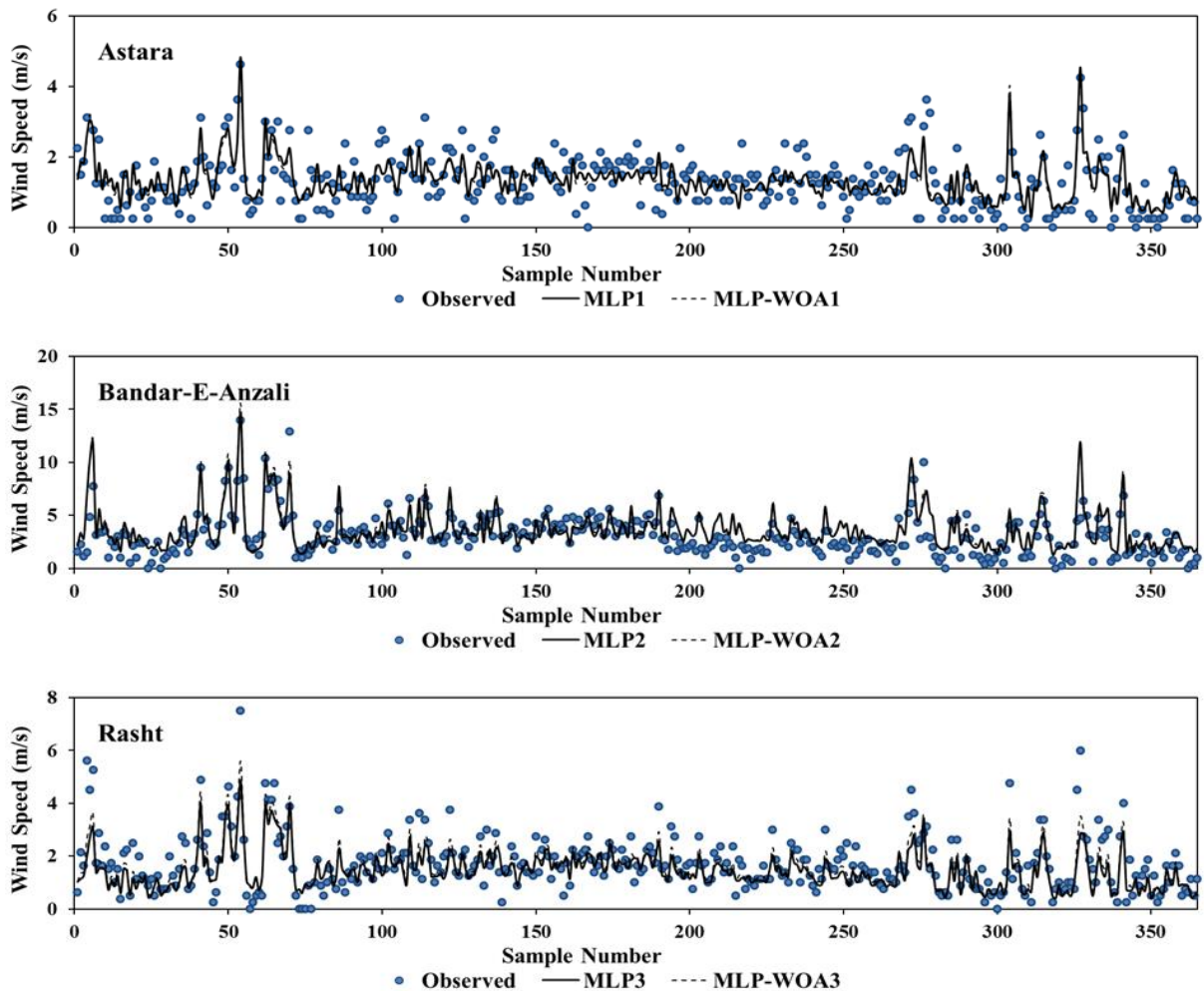

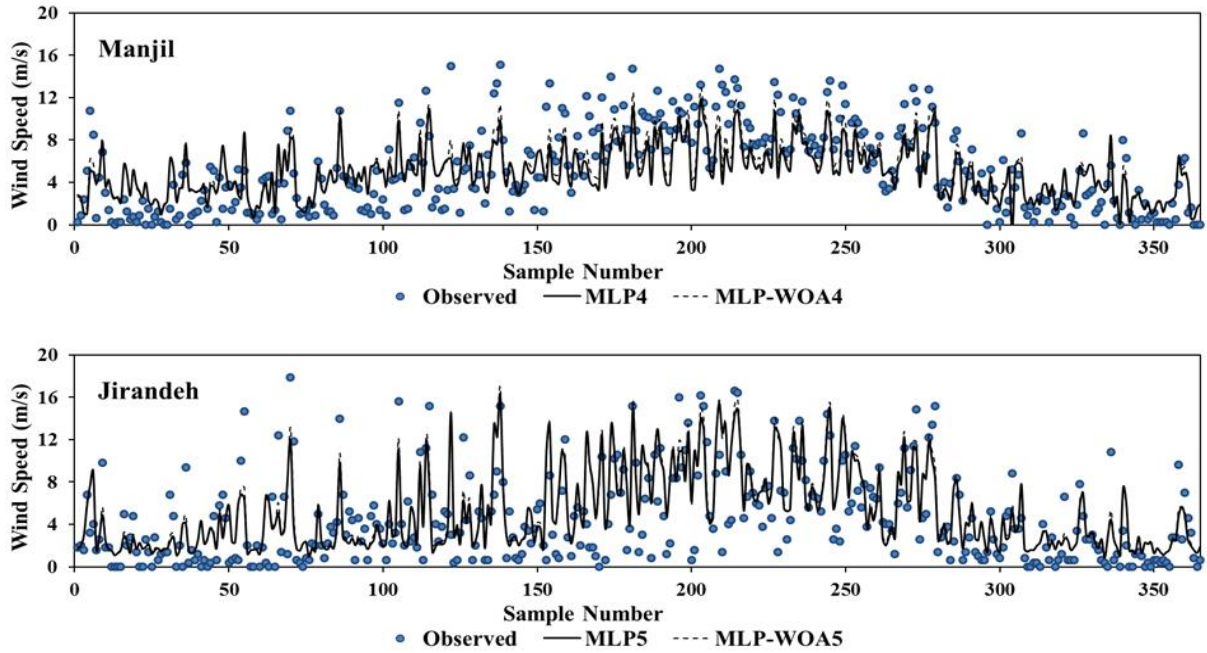

**Fig. 6.** Comparison of the predicted and observed daily wind speed values using the hybrid MLP-WOA and classical MLP models

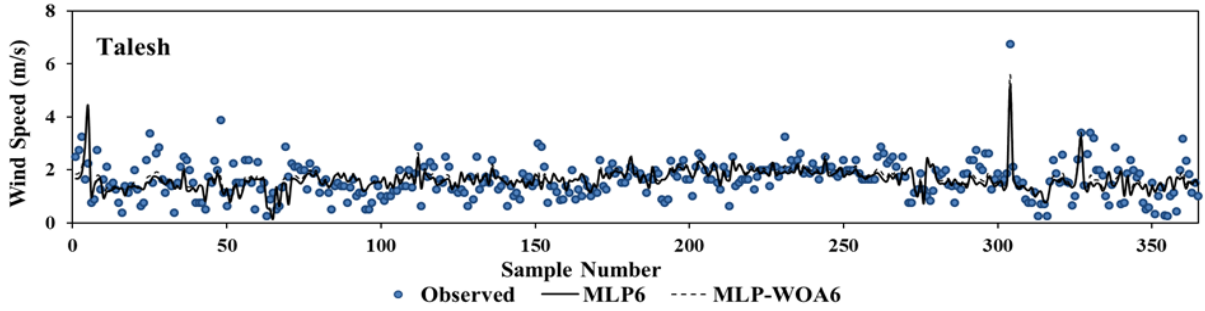
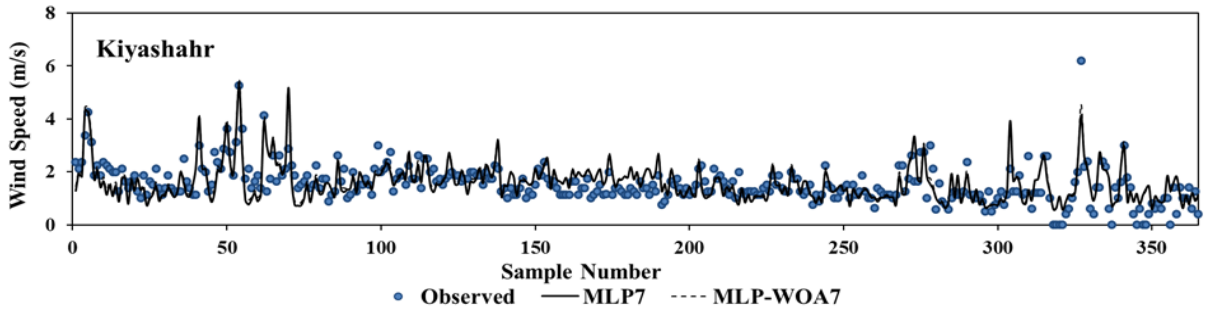
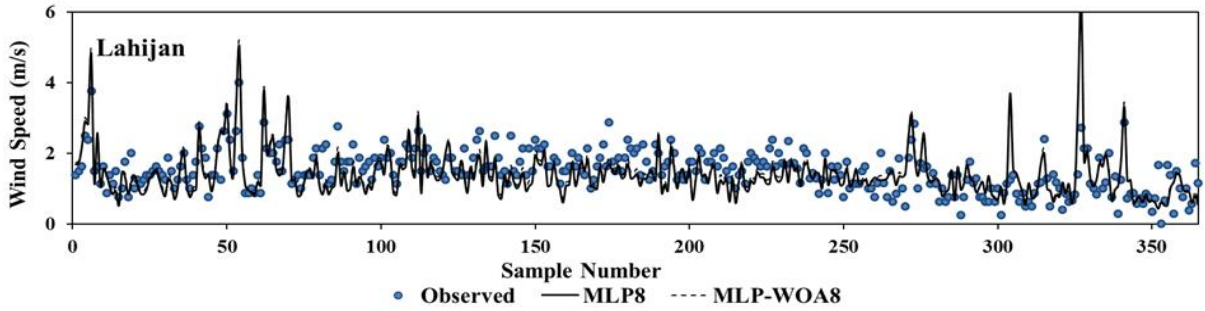
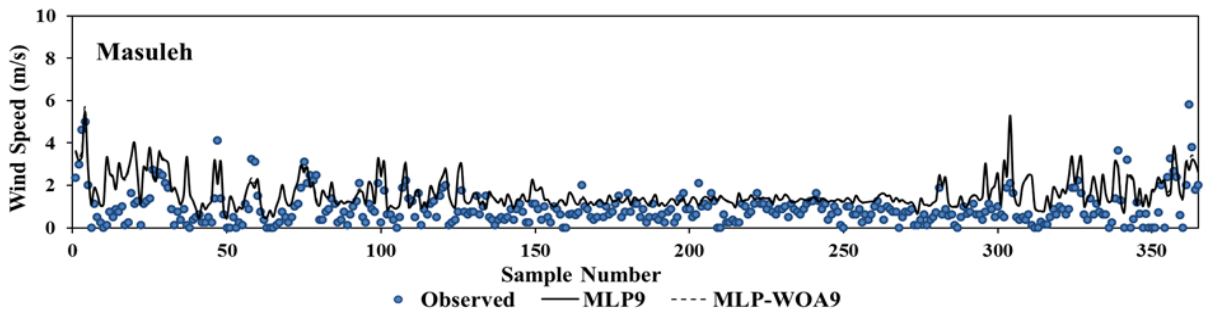
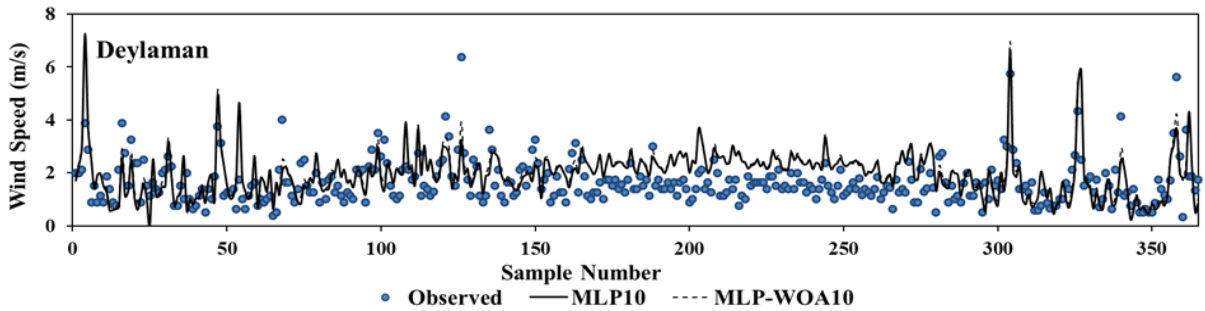

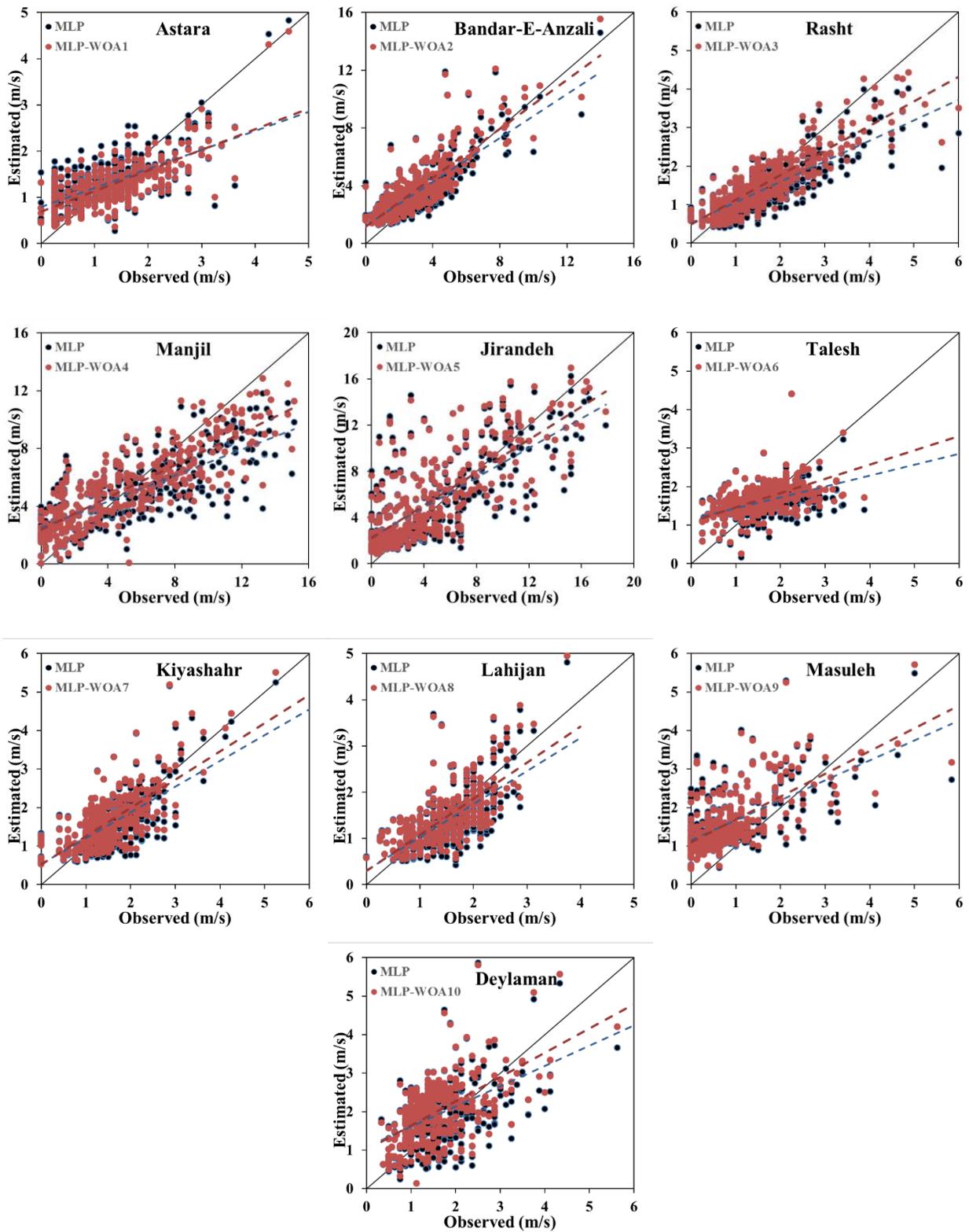

**Fig. 6.** Continue

**Fig. 7.** Scatter plots of estimated and observed values of wind speed at various stations

## 4. Conclusion

One of the problems of artificial intelligence algorithms is selecting finest weights in the layers of neural networks that must permit the extraction of the relevant features within the input information for creating an accurate model. Constructing the best predictive model demands input data which is considered as a crucial and useful tool for calculation of wind energy potential. In the present study, the utility of a reliable and powerful method for predicting the wind speed for ten locations is revealed, where the wind speed amount of the target location was forecasted using input data of neighboring reference locations. In the current study by using the MLP and MLP-WOA models where the Whale Optimization algorithm combined with standalone MLP for each of the ten target station, daily wind speed values are predicted. Furthermore, other climate or atmospheric information is not used for wind speed prediction with this method. In order to evaluate the performance of MLP-WOA, Several statistical indices were used. The results demonstrated that the hybrid MLP-WOA model has high accuracy in the estimation of wind speed almost in all of the stations.

**Acknowledgments**

The support of the University of Tabriz Research Affairs Office is acknowledged.